\documentclass{article}
\usepackage[T1]{fontenc}
\usepackage{url}
\usepackage{amsmath}
\usepackage{amssymb}
\usepackage{amsfonts}
\usepackage{xcolor}
\usepackage[left=2cm,right=2cm,top=2cm,bottom=2cm]{geometry}
\usepackage{float}
\usepackage{graphicx}
\usepackage{centernot} 
\usepackage{enumitem}
\usepackage{comment}
\usepackage{subcaption}
 \newtheorem{Definition}{Definition}
 \newtheorem{Theorem}{Theorem}

\def\bw{\mathbf w}

\def\+{\hspace{-1mm}+\hspace{-1mm}}
\def\={\hspace{-3mm}=\hspace{-3mm}}

\usepackage[vlined,ruled,linesnumbered]{algorithm2e}
\usepackage{gensymb}
\usepackage{tikz}
\usepackage{fixltx2e}
\usepackage[noend]{algpseudocode}

\usepackage[vlined,ruled,linesnumbered]{algorithm2e}
\SetKwInOut{Input}{input}
\SetKwInOut{Output}{output}
\SetKwComment{Comment}{}{}

\SetCommentSty{mycmtsty}

\title{Can neural networks extrapolate? \\ Discussion of a theorem by Pedro Domingos}
\author{Adrien Courtois, Jean-Michel Morel, Pablo Arias}
\date{ENS Paris-Saclay}

\begin{document}

\maketitle




\paragraph{Abstract}
Neural networks trained on large datasets by minimizing a loss have become the state-of-the-art approach for resolving data science problems, particularly  in computer vision, image processing and natural language processing. In spite of their striking results, our theoretical understanding about how neural networks operate is limited. In particular, what are the interpolation capabilities of trained neural networks?
In this paper we discuss a theorem of Domingos stating that ``every machine learned by continuous gradient descent is approximately a kernel machine''.  According to Domingos, this fact leads to conclude that all machines trained on data are mere kernel machines. We first extend Domingo's result in the discrete case and to networks with vector-valued output. We then study its relevance and significance on simple examples. We find that in simple cases, the ``neural tangent kernel'' arising in Domingos' theorem does provide understanding of the networks' predictions. Furthermore, when the task given to the network grows in complexity, the interpolation capability  of the network can be effectively explained by Domingos' theorem, and therefore is limited. We illustrate this fact on a classic perception theory problem: recovering a shape from its boundary.

 
\section{Introduction}

 Artificial neural networks (NNs) are complex non-linear functions $\mathcal{N}(w,x)$ obtained by combining multiple simple units, in a structure that is reminiscent of how neurons are organized in the brain.
NNs are parameterized by a vector $w$ of millions or billions of parameters. These parameters are optimized on very large datasets of input-output pairs $(x_n, y_n)$ to realize a single complex task, ranging from
image classification to natural language processing. Training amounts to setting the parameters of the neural network $\mathcal{N}(w;x)$ via the minimization of the performance of an average objective function $\sum_n L(\mathcal{N}(x_n; w),y_n)$ of the network for a given task over a large dataset.

In spite of striking practical successes, very little is understood about NNs. Modern NNs stack many layers of computations making it very hard to interpret the operations ``learned'' by the network during training.

Our purpose here is to  discuss a theorem  of Pedro Domingos stating that ``every machine learned by gradient descent is approximately a kernel machine'' \cite{domingos2020every},
thus allowing to interpret NNs within the framework of kernel methods, which have a well-developed theory \cite{hofmann2008kernel,smola1998learning}.
Kernel machines  $ x \mapsto \sum_{n=1}^N \alpha_k K(x, x_n) + \alpha_0 $ work by comparing a new sample $x$ with the ones $x_n$ belonging to the training set through a kernel function $K(x,x_n)$. This simple one to one comparison structure brings insight  on the  interpolation capabilities  of such methods, and a direct interpretation of each prediction. Using the kernel that naturally arises in Domingos' theorem brings the hope of   better understanding the predictions of a neural network.

In a way, all neural networks trained under supervision can be seen as mere interpolators. Domingo's theorem provides a structure to this interpolator.
In this work, we question its capabilities.
In particular, in the case where there exists an oracle $f$ such that $y_n = f(x_n)$ for every pairs $(x_n, y_n)$, we ask ourselves if a network can converge to this oracle when the network's structure allows it.
Such cases have been partially observed in the literature \cite{newson2020processing}.
This question can be perceived as a question of both ``interpolation" and ``extrapolation", but is not to be confused with the notion of ``generalization" which has statistical connotations \cite{bartlett2002rademacher, bartlett2017spectrally}.

Although trained neural networks are often considered as black boxes due to their lack of interpretability, Domingos' theorem suggests a way of interpreting their predictions. Indeed, kernel machines provide us with two tools for interpretation: the \textit{kernel function} and the \textit{feature vector}. The feature vector
is defined by Mercer's theorem, which indicates that a comparison of two samples $x, x'$ using a nonnegative kernel $K(x,x')$ amounts to computing a scalar product $K(x,x') = \langle\phi(x), \phi(x')\rangle$ in a certain feature vector $\phi(x)$. We shall use these two tools to derive an understanding of how a network makes its predictions in simple, low-dimensional learning examples. We discover that in these simple cases, the associated kernel function and feature vector  shed light on the inner workings of the neural network.

It is well known that for complex problems, neural networks fail to extrapolate to data that are too different from the ones seen during training. This is precisely the point made by Domingos in his theorem, which suggests that neural networks behave as kernel machines and will therefore interpolate correctly near their training domain, but have no reason  to extrapolate well on samples that stand farther away from this domain.
We shall analyze this fact on datasets of gradually increasing complexity. We'll find that in the simplest cases, neural networks interpolate perfectly, but for more complex tasks, this stops and the network is unable to extrapolate to samples that are too different from the ones in the training set.

The kernel highlighted by Domingos in his theore is similar to the Neural Tangent Kernel (NTK) \cite{jacot2018neural}, a kernel that is used for interpreting neural networks in the asymptotic case where the considered networks have an infinite width. Domingos' kernel is more general than the NTK, as it can be computed on any neural network. Nonetheless, due to their formal similarity, we will refer to Domingos' kernel as a neural tangent kernel (NTK). 




Our plan follows. 
\begin{itemize}
    \item In Section \ref{neuralnetworksection} we  recall the basics of neural networks, kernel machines, and of their training.   
    \item In Section \ref{sec:domingo-proof}, we give the necessary notation and an analysis of Domingo's proof. Notably, we propose an alternative formulation of Domingos' result that is perhaps more self-explanatory;
    \item we extend Domingos' proof to the discrete setting (Section \ref{sec:discrete-domingo-proof}) and the multi-dimensional case (Section \ref{sec:domingo-multidim});
    \item we mathematically study the implication of these theorems in the case of a linear regression (Section \ref{sec:specialcase-linearreg}). We also remark the linear separability of the feature space induced by the NTK (Section \ref{sec:theo-svm});
    \item we numerically study the theorems and their limitations in simple cases where the network solves the task perfectly, and use the feature space and the NTK to gain understanding about the network's predictions (Section \ref{sec:numverif-domingo});
    \item we further study this theorem with an application where we teach a network to learn connectivity of planar shapes (Section \ref{sec:shape-completion-2d}) and find limitations in interpolation capabilities and interpretability.
\end{itemize}

\section{Neural networks and kernel machines} \label{neuralnetworksection}

\subsection{Neural Networks}
A neural network $\mathcal{N}$ can be described as a family of functions $(f_i)_{i \in [|1,D|]}$ and a vector of weights $w \in \mathbb{R}^d$ such that the network is defined as
$$\mathcal{N}(x; w) = f_D(f_{D-1}(\hdots (f_2(f_1(x; w); w) \hdots; w); w).$$
Notably, each function $f_i$ is differentiable almost everywhere and is parameterized by weights stored in $w \in \mathbb{R}^d$. Most often, these functions are of the following form:
\begin{itemize}
    \item[-] \textit{Fully-connected affine:} The function operates a linear transformation of the input: $f_i(x) = W x + b$ where $W$ is a matrix and $b$ in a vector.
    \item[-] \textit{Convolutional:} Instead of a matrix multiplication, a convolution can be applied instead.
    \item[-] \textit{Activation function:} To mimic the behavior of the brain where a neuron can only transmit if a certain voltage is reached, non-linear functions are used and applied element-wise. For instance the ReLU \cite{agarap2018deep} $f_i(x) = \max(0, x)$.
    \item[-] \textit{Down-sampling:} The spatial resolution of the input is reduced by a factor at least two.
    \item[-] \textit{Up-sampling:} The spatial resolution of the input is increased by a factor at least two using an interpolation algorithm.
    \item[-] \textit{Normalization:} The input to the function is normalized according to some pre-recorded statistics to avoid for large change in statistics from one training iteration to the next.
\end{itemize}

The choice of a network architecture \textit{i.e.} which function to use and where, does not have a clear answer as of yet. Most researches use heuristics or draw inspiration from existing algorithms that are known to work well.

\label{sec:dl:cnn-overview}



\subsubsection{Training a neural network}
\label{sec:dl:training}

Neural networks are trained using large number of examples pairs $(x,y)$ of inputs and the desired outputs (i.e. noisy  and noiseless image pairs) stored in a dataset $\mathcal{D} := (x_n, y_n)_{n \in [|1,N|]} \in \left(\mathbb{R}^p \times \mathbb{R}^q\right)^N$.  The desired output $y$ is often called the \emph{label}, a term inherited from the application of machine learning to classification. 

The objective of training is to minimize the risk, which is the expected value of the loss 
\begin{equation}
\mathcal{L}(w) = \int L(\mathcal{N}(x; w),y) \mathrm{d}\mathbb{P}(x,y), \label{expectation}
\end{equation}
where $\mathbb{P}$ is the probability distribution of the data, and $w$ denotes the vector of all parameters of the network. The distribution $\mathbb{P}$ is usually unknown and  the empirical distribution $\mathbb{P} = \frac{1}{N} \sum_{n=1}^N \delta_{(x_n, y_n)}$ is used instead, so that
\begin{equation}
	\mathcal{L}(w) = \frac{1}{N}\sum_{n = 1}^N L(\mathcal{N} (x_n; w), y_n).
\end{equation}
%
%
To minimize this risk - or this loss - a gradient descent algorithm is used, which therefore reaches a local minimum, as the functional is generally not convex.


The gradient $\nabla_w  \mathcal{L}$ of the loss over the training data is computed with respect to the set of parameters $w$. 
In practice, due to computational limitations, the stochastic gradient descent is used instead.
The stochastic component consists in randomly sampling a batch of samples at each iteration and computing the gradient with respect to those samples.
This random subset of samples is called a \textit{mini-batch}. 
%

\begin{algorithm}[H]
 \While{\text{stopping criterion not met}}{
  Sample mini-batch of $m$ samples $x_1, x_2, ..., x_m$ and corresponding targets $y_1, y_2, \hdots, y_m$\;
  Compute gradient estimate: $\Delta w \longleftarrow \frac{1}{m} \nabla_w  \sum_{n=1}^m L(\mathcal N(x_n; w), y_n)$

	Update the parameters: $w \longleftarrow w - \eta \cdot \Delta w$
 }
 \caption{Stochastic gradient descent.} 
\end{algorithm}
The length of the gradient step is controlled by $\eta \in \mathbb{R}^{+}_*$, the \textit{learning rate}, which can be variable. More sophisticated methods which allow for faster convergence, such as Adam, RMS prop, or SGD with Nesterov momentum are usually applied~\cite{kingma2014adam, nesterov1983method}.


\color{black}



\subsection{Kernel Machines}

Kernel machines are simple and effective mathematical models used for machine learning. They proceed by  comparing a given input sample $x$ to each training sample $x_n$ using a \textit{kernel} function $x \in \mathbb{R}^p \mapsto K(x,x_n) \in \mathbb{R}$,  and obtaining a value for $x$ with weights depending on the kernel. The simplest example of kernel is the so-called linear kernel $K(x, x') = \langle x, x' \rangle_{\mathbb{R}^p}$ \cite{slidesmvakernelmethods}.

\medskip 
Given a dataset $\mathcal{D} := (x_n, y_n)_{n \in [|1,N|]} \subset \left(\mathbb{R}^p \times \mathbb{R}\right)^N$, a kernel machine is defined by a kernel $K: \mathbb{R}^p \times \mathbb{R}^p \to \mathbb{R}$ and a set of coefficients $(\alpha_n)_{n \in [|0,N|]}$, leading to define a decision function $f$ by
\begin{equation}
    \label{eq:kernel-machine}
    \begin{aligned}
        f \colon  & \mathbb{R}^p \to \mathbb{R} \\
        & x \mapsto \sum_{n=1}^N \alpha_k K(x, x_n) + \alpha_0.
    \end{aligned}
\end{equation}

\medskip 
By Mercer's theorem \cite{mercer1909xvi}, if $(K(x_i, x_j))_{(i,j) \in [|1,N|]^2}$ is positive semi-definite, there is a function $\phi: \mathbb{R}^p \to \mathbb{R}^s$ such that $K(x, \tilde{x}) = \langle \phi(x), \phi(\tilde{x}) \rangle$ for all $x, \tilde{x} \in \mathbb{R}^p$, where $s$ can be arbitrarily large. Therefore, the kernel trick amounts to projecting the data into an arbitrary high-dimensional space  (referred to as feature space) where the comparison between samples is more meaningful.

\medskip
In most kernel machine applications, the function $K$ is manually chosen beforehand based on the properties of the manifold containing the dataset. The parameter $\boldsymbol{\alpha}$ is set afterwards using an optimization algorithm.

\medskip
A classic example of kernel $K$ is the RBF kernel \cite{broomhead1988radial}  $K(x, \tilde{x}) = \mathcal{G}(d(x, \tilde{x}))$, where $\mathcal{G}$ is a Gaussian and   $d$  an arbitrary distance. In that case, the kernel directly defines a similarity measure \textit{i.e.} the function 
reaches its maximal value when $\tilde x = x$. This makes for the interpretability of this method:   the closest  samples are those responsible for the response of the model to a given input.

\section{Domingos' theorem, proof and variants, interpretation} \label{sec:domingo-proof}

\subsection*{Notation}

We consider a learning data set $\mathcal{D} := (x_n, y_n)_{n \in [|1,N|]} \in \left(\mathbb{R}^p \times \mathbb{R}\right)^N$ and a parametric machine learning device defined by
\begin{align*}
  \mathcal{N} \colon \mathbb{R}^p \times \mathbb{R}^d &\to \mathbb{R} \\
  (x,w) &\mapsto \mathcal{N}(x;w_1, \hdots, w_j, \hdots, w_d) = \mathcal{N}(x;w)
\end{align*}
where $x \in \mathbb{R}^p$ corresponds to an input and $w \in \mathbb{R}^d$  is the  vector of all parameters of the learning device.
For $n \in [|1,N|]$, we define the prediction $\hat{y}_n (w)$ of the model for a learning data point $x_n$ by
\begin{align*}
  w &\mapsto \hat{y}_n = \mathcal{N}(x_n;w_1, \hdots, w_j, \hdots, w_d) = \mathcal{N}(x_n;w).
\end{align*}
The learning process  minimizes a loss  $\mathcal{L}$ defined by
\begin{align*}
    \mathcal{L} \colon \mathbb{R}^d & \to \mathbb{R} \\
    w &\mapsto \mathcal{L}(w) = \mathcal{L}(w_1, \hdots, w_j, \hdots, w_d):= \frac{1}{N} \sum_{n=1}^N L(\hat{y}_n(w), y_n),
\end{align*}
where $L: \mathbb{R}^2 \to \mathbb{R}$ is a given loss function such as $L(\hat{y}, y)=(\hat{y} - y)^2$. This criterion  is minimized by standard gradient descent 
\begin{equation}
    \label{eq:dyn_grad_desc}
    \frac{dw(t)}{dt} = -\nabla_w \mathcal{L}(w(t)).
\end{equation}
This requires some regularity for $\mathcal{N}(x;w)$ with respect to $w$. For neural networks, this regularity is ensured if the activation function is smooth.  With exception of the frequently used ReLU activation \cite{agarap2018deep} $f(x)=\max(0,x)$, all current activation functions are $C^\infty$, including several smooth versions of the ReLU called respectively GELU \cite{hendrycks2016gaussian}, ELU \cite{clevert2015fast}, SELU \cite{klambauer2017self} and Swish \cite{ramachandran2017searching} (see \cite{dubey2021comprehensive} for a comparison). We shall limit our analysis to such activation functions, which guarantee that the learning machine is at least $C^2$, and generally $C^\infty$.
We shall denote by $\bw: t \in \mathbb{R}_+ \mapsto w(t) = (w_1(t), \hdots, w_j(t), \hdots, w_d(t)) \in \mathbb{R}^d$ the solution of the above equation for a given initial set of parameters $w(0) = w_0 \in \mathbb{R}^d$.  We call $\bw$ a \textit{learning path}.

\begin{Definition}
The {\rm tangent kernel} of a $C^2$ learning function $\mathcal{N}$ at parameter $w$ is defined as\vspace{-2mm} $$\forall (x, \tilde{x}) \in \left(\mathbb{R}^p\right)^2, K_{\mathcal{N},w}(x, \tilde{x})= \langle \nabla_w \mathcal{N}(x; w), \nabla_w \mathcal{N}(\tilde{x};w)\rangle_{\mathbb{R}^d}.$$
\end{Definition}

\begin{Definition}\label{def:feature-space}
We call {\rm feature vector} $\phi(x)$ of an input $x\in \mathbb{R}^p$  induced by the {\rm tangent kernel} of a $C^2$ learning function $\mathcal{N}$ at parameter $w$, the vector
$$\phi(x) := \nabla_w \mathcal{N}(x; w) \in \mathbb{R}^d.$$
\end{Definition}

\begin{Definition}
The {\rm path kernel} associated with a $C^2$ learning function $\mathcal{N}$ and a learning path $\bw$ up to time $t$ is
\vspace{-3mm}
$$\forall t \in \mathbb{R}_+, \forall (x, \tilde{x}) \in \left(\mathbb{R}^p\right)^2, K_{\mathcal{N},\bw}(x, \tilde{x}; t) = \int_0^t \langle \nabla_w \mathcal{N}(x; w(s)), \nabla_w \mathcal{N}(\tilde{x}; w(s)) \rangle_{\mathbb{R}^d} ds.$$
\end{Definition}

\begin{Theorem} (Domingos \cite{domingos2020every})
Consider a twice continuously differentiable learning machine model $\mathcal{N}$ and $\bw$ the path of parameters learned from a training set $\mathcal{D}$ by gradient descent of a loss function $\mathcal{L}$.
Then $\mathcal{N}$ satisfies
$$\forall t \in \mathbb{R}_+, \forall x \in \mathbb{R}^p, \mathcal{N}(x;w(t)) = \sum_{n=1}^N a_n(x) K_{\mathcal{N},\bw}(x, x_n; t) + b(x),$$
where $K_{f,\bw}$ is the  path kernel associated with $\mathcal{N}$ and the learning path $\bw$ taken during gradient descent, $a_n$ is the average value of $\frac{\partial L}{\partial \hat{y}_n}$ along the path weighted by the tangent kernel, and $b=\mathcal{N}(\ \cdot \ ;w(0))$ is the initial model.  \label{Theorem1}
\end{Theorem}

\noindent {\bf Proof}
By the chain rule,

$$\frac{\partial \mathcal{N}(x;w(t))}{\partial t} = \sum_{j=1}^d \frac{\partial \mathcal{N}}{\partial w_j}(x ; w(t)) \frac{d w_j}{dt}(t),$$
where $w : t \in \mathbb{R}_+ \mapsto (w_j(t))_{j \in [|1,d|]} \in \mathbb{R}^d$. Using Equation \eqref{eq:dyn_grad_desc}, this yields

$$\frac{\partial \mathcal{N}(x;w(t))}{\partial t}=\sum_{j=1}^d \frac{\partial \mathcal{N}}{\partial w_j}(x; w(t)) \left(-\frac{\partial \mathcal{L}}{\partial w_j}(w(t))\right).$$
But  $\mathcal{L}(w)=\frac{1}{N}\sum_{i=1}^N L(\hat{y}_n(w), y_n) = \frac{1}{N} \sum_{n=1}^N L(\mathcal{N}(x_n; w), y_n)$, and using again the chain rule, 

$$\frac{\partial \mathcal{N}(x; w(t))}{\partial t}=\sum_{j=1}^d \frac{\partial \mathcal{N}}{\partial w_j}(x; w(t))\left(-\frac{1}{N}\sum_{n=1}^N\frac{\partial L}{\partial \hat{y}}(\hat{y}_n(w(t)), y_n)\frac{\partial \mathcal{N}}{\partial w_j}(x_n ; w(t))\right).$$
Rearranging terms:
$$\frac{\partial \mathcal{N}(x; w(t))}{\partial t}=-\frac{1}{N}\sum_{n=1}^N\frac{\partial L}{\partial \hat{y}}(\hat{y}_n(w(t)), y_n) \sum_{j=1}^d \frac{\partial \mathcal{N}}{\partial w_j}(x; w(t))\frac{\partial \mathcal{N}}{\partial w_j}(x_n; w(t)).$$
Let $L'(\hat{y}(w(t))_n, y_n) = \frac{\partial L}{\partial \hat{y}}(\hat{y}_n(w(t)), y_n)$. Then using the definition of the tangent kernel 
$$\forall (x, \tilde{x}) \in \left(\mathbb{R}^p\right)^2, K_{\mathcal{N},w}(x, \tilde{x})= \langle \nabla_w \mathcal{N}(x; w), \nabla_w \mathcal{N}(\tilde{x}; w) \rangle = \sum_{j=1}^d\frac{\partial \mathcal{N}}{\partial w_j}(x;w) \frac{\partial \mathcal{N}}{\partial w_j}(\tilde{x};w),$$
\begin{equation}
    \label{eq:dyn_ed_continuous_kernel}
    \frac{\partial  \mathcal{N}(x; w(t))}{\partial t}=-\frac{1}{N}\sum_{n=1}^N L'(\hat{y}_n(w(t)), y_n) K_{\mathcal{N},w(t)}(x, x_n).
\end{equation}
Integrating between 0 and $t$ yields
\begin{equation}
    \label{eq:ntk_interp1}
    \mathcal{N}(x; w(t)) = \mathcal{N}(x; w(0)) - \frac{1}{N} \int_0^t \sum_{n=1}^N L'(\hat{y}_n(w(s)), y_n) K_{\mathcal{N}, w(s)}(x, x_n) ds.
\end{equation}
Multiplying and dividing each term in the sum by $\int_0^t K_{\mathcal{N},w(s)}(x,x_n)ds,$ we get

$$\mathcal{N}(x; w(t)) = \mathcal{N}(x; w(0)) - \frac{1}{N} \sum_{n=1}^N \left(\frac{\int_0^t L'(\hat{y}_n(w(s)), y_n) K_{\mathcal{N},w(s)}(x, x_n) ds}{\int_0^t K_{\mathcal{N},w(s)}(x,x_n) ds} \right) \int_0^t K_{\mathcal{N},w(s)}(x,x_n) ds,$$ which yields by definition of the path tangent kernel
$$\mathcal{N}(x;w(t)) = \mathcal{N}(x; w(0)) - \frac{1}{N} \sum_{n=1}^N \bar{L'}(x, x_n, \hat{y}_n, y_n; t) K_{f,\bw}(x,x_n; t),$$
where
$$\bar{L}'(x, x_n, \hat{y}_n, y_n; t) = \frac{\int_0^t L'(\hat{y}_n(w(s)), y_n) K_{\mathcal{N}, w(s)}(x, x_n) ds}{\int_0^t K_{\mathcal{N}, w(s)}(x, x_n) ds},$$
hence
$$\mathcal{N}(x; w(t))= \sum_{n=1}^N a_n(x;t) K_{\mathcal{N},\bw}(x,x_n; t) + b(x),$$ where $K_{\mathcal{N},\bw}(x,x_n;t)$ is the path kernel associated with the learning process up to time $t$, $b$ is the initial learning function, $a_n = -\frac{1}{N}\bar{L'}(x,x_n,\hat{y}_n, y_n;t)$ is the loss derivative weighted by the tangent kernel (note that these weights are not necessarily positive). This concludes the proof of Theorem 1.

\subsubsection{Discussion}
Even if, formally, the result of Theorem \ref{Theorem1} has   the aspect of a kernel machine, the dependence of $a_n(x;t)$ on $x$ is difficult to interpret. We can reach a more accessible interpretation in the line of Domingos by stopping the reasoning at Equation \eqref{eq:ntk_interp1}. We get, for the standard choice for the loss $L(\hat{y}, y) = \frac{1}{2} (\hat{y} - y)^2$:
\begin{equation}
    \label{eq:interp_continuous}
    \mathcal{N}(x;w(t))= \mathcal{N}(x; w(0)) - \frac{1}{N}\sum_{n=1}^N \int_0^t (\mathcal{N}(x_n; w(s)) - y_n) \cdot \langle \nabla_w \mathcal{N}(x; w(s)), \nabla_w \mathcal{N}(x_n; w(s)) \rangle_{\mathbb{R}^d} ds.
\end{equation}
Theorem 1 and Equation \eqref{eq:interp_continuous} inform us about the behavior of a trained neural network (or of any learning machine using gradient descent). This formula indicates that once trained, a neural network  will be ``comparing'' a given new  sample $x$ to all samples $x_n$ in the training set. Indeed, we can rewrite Equation \eqref{eq:interp_continuous} as
$$\mathcal{N}(x;w(t)) = b(x) - \frac{1}{N}\sum_{n=1}^N \int_{0}^t a_n(s) \cdot K_{\mathcal{N}, w(s)}(x, x_n) ds,$$
which yields a simple  formulation that is almost identical to a standard kernel machine (Equation \eqref{eq:kernel-machine}). The exception lies in the bias term $b(x)$ depending on the input sample $x$, and in the kernel being integrated over time.

\medskip \noindent 
What we actually get from Equation \eqref{eq:dyn_ed_continuous_kernel} is that the time derivative of the network is exactly a kernel machine. More generally, this conclusion applies to any learning machine trained by gradient descent.

\subsection{Extension of Domingo's theorem to the discrete setting} \label{sec:discrete-domingo-proof}
In practice, neural networks are not learned using the dynamic described in Equation \ref{eq:dyn_grad_desc} which implies using an infinite number of training steps with an infinitesimal step size. Instead, neural networks are trained for finite number of steps and using a relatively high step; typically in the range $[10^{-5}, 10^{-1}]$. In this situation, the dynamic of gradient descent for $K$ steps is given by
\begin{equation}
    \label{eq:discrete_dyn_grad_desc}
    \forall k \in [|0,K-1|], w(k+1) = w(k) - \eta_k \nabla_w \mathcal{L}(w(k)),
\end{equation}
where $(\eta_k)_{k \in [|0,K-1|]}$ is the sequence of steps considered. Then the learning path $\bw$ associated with an initial condition $w(0) = w_0 \in \mathbb{R}^d$ is defined as a discrete path $\bw: k \in [|0, K|] \mapsto w(k) = (w_1(k), \hdots, w_j(k), \hdots, w_d(k)) \in \mathbb{R}^d$.

\noindent
These considerations lead us  to consider a discrete formulation of Theorem 1.

\begin{Definition}
The discrete neural tangent kernel of a 
$C^2$ learning function $\mathcal{N}$ is defined for each given iteration number $k \in [|0, K|]$, by
$$\forall (x, \tilde{x}) \in \left(\mathbb{R}^p\right)^2, K_{\text{NTK}}(x,\tilde{x};k) = \left\langle \nabla_w \mathcal{N}(x; w(k)), \nabla_w \mathcal{N}(\tilde{x}; w(k)) \right\rangle_{\mathbb{R}^d}.$$
\end{Definition}

\begin{Theorem}
Consider a $C^2$  learning machine model $\mathcal{N}$ and $\bw$ the discrete path of parameters learned from a training set $\mathcal{D}$ by discrete gradient descent with steps $(\eta_k)_{k \in [|0,K-1|]}$ of a loss function $\mathcal{L}$.  Assume that the gradient of the loss $L$ Lipschitz-continuous. Then $\mathcal{N}$ can be expressed as
\begin{equation}
    \label{eq:domingo_discrete}
    \forall x \in \mathbb{R}^p, \mathcal{N}(x; w(K)) = \mathcal{N}(x; w(0)) - \frac{1}{N} \sum_{n=1}^N \sum_{k=0}^{K-1} \eta_k \frac{\partial L}{\partial \hat{y}} (\hat{y}_n(w(k)), y_n) K_{\text{NTK}}(x, x_n; k) + O\left(\sum_{k=0}^{K-1} \eta_k^2\right),
\end{equation}
where $\hat{y}(k)_n = \mathcal{N}(x_n; w(k))$.
\end{Theorem}

\noindent {\bf Proof}

\noindent 
The Taylor approximation of $w \in \mathbb{R}^d \mapsto \mathcal{N}(x ; w) \in \mathbb{R}$ for a fixed $x \in \mathbb{R}^p$ gives 
$$\mathcal{N}(x;w(k+1)) = \mathcal{N}(x;w(k)) + \left\langle \nabla_w \mathcal{N}(x;w(k)),  w(k+1)-w(k) \right\rangle_{\mathbb{R}^p} + O(\|w(k+1)-w(k)\|_2^2).$$
Combining this equation with Equation \eqref{eq:discrete_dyn_grad_desc} yields
$$\mathcal{N}(x; w(k+1)) = \mathcal{N}(x;w(k)) - \eta_k \frac{1}{N} \sum_{n=1}^N \left\langle \nabla_w \mathcal{N}(x; w(k)), \nabla_w \mathcal{L} (w(k)) \right\rangle_{\mathbb{R}^p} + O(\eta^2_k).$$
 By the same chain rule trick as for the continuous case we rewrite one of the above gradients as
$$\frac{\partial \mathcal{L}}{\partial w_j} (w(k)) = \frac{1}{N} \sum_{n=1}^N \frac{\partial \mathcal{N}}{\partial w_j} (x_n ; w(k)) \frac{\partial L}{\partial \hat{y}}(\hat{y}_n(w(k)), y_n).$$
Injecting this equation into the previous one yields
\begin{align*}
    \mathcal{N}(x; w(k+1)) &= \mathcal{N}(x; w(k)) - \eta_k \frac{1}{N} \sum_{n=1}^N \frac{\partial L}{\partial \hat{y}}(\hat{y}_n(w(k)), y_n) \left\langle \nabla_w \mathcal{N}(x; w(k)), \nabla_w \mathcal{N}(x_n; w(k)) \right\rangle_{\mathbb{R}^p} + O(\eta^2_k) \\
    &= \mathcal{N}(x; w(k)) - \eta_k \frac{1}{N} \sum_{n=1}^N \frac{\partial L}{\partial \hat{y}}(\hat{y}_n(w(k)), y_n) K_{\text{NTK}}(x, x_n; k) + O(\eta^2_k).
\end{align*}
As for in the continuous case, a telescopic summation results in
$$\mathcal{N}(x; w(K)) = \mathcal{N}(x; w(0)) - \frac{1}{N} \sum_{n=1}^N \sum_{k=0}^{K-1} \eta_k \frac{\partial L}{\partial \hat{y}}(\hat{y}_n(w(k)), y_n) K_{\text{NTK}}(x, x_n; k) + O\left(\sum_{k=0}^{K-1} \eta_k^2\right).$$

\subsection{Extension of Domingo's theorem to the multi-dimensional output case} \label{sec:domingo-multidim}

Most of the time, neural networks feature multidimensional outputs.
In this section we extend the theorem to cope with this more general setup.
Consider a training data set $\mathcal{D} = (x_n, y_n)_{n
\in [|1,N|]} \in \left(\mathbb{R}^p \times \mathbb{R}^q\right)^N$ with a
learning function
\begin{align*}
    \mathcal{N} \colon \mathbb{R}^p \times \mathbb{R}^d &\to \mathbb{R}^q \\
    (x, w) &\mapsto \mathcal{N}(x; w) = \mathcal{N}(x; w_1, \hdots, w_j, \hdots, w_d) := (\mathcal{N}_1(x;w), \hdots, \mathcal{N}_q(x;w)).
\end{align*}
Like in the one-dimensional case, we denote the  prediction at a data point $x_n$ of the  network with parameter $w$  by
\begin{align*}
    \forall m \in [|1,q|], \forall n \in [|1,N|], \hat{y}_n^{(m)} \colon \mathbb{R}^d & \to \mathbb{R} \\
    w &\mapsto \hat{y}^{(m)}_n(w) = \hat{y}^{(m)}_n(w_1, \hdots, w_j, \hdots, w_d) = \mathcal{N}_m(x_n; w). 
\end{align*}
Like in the one-dimensional case we assume  the gradient descent \eqref{eq:discrete_dyn_grad_desc} and adopt any loss 

\begin{align*}
    L \colon \mathbb{R}^q \times \mathbb{R}^q &\to \mathbb{R} \\
    (\hat{y}, y) &\mapsto L(\hat{y}, y) := L(\hat{y}^{(1)}, \hdots, \hat{y}^{(j)}, \hdots, \hat{y}^{(q)}, y^{(1)}, \hdots, y^{(j)}, \hdots, y^{(q)}).
\end{align*}
For instance, one could consider $L(\hat{y}, y) = L(\hat{y}^{(1)}, \hdots, \hat{y}^{(j)}, \hdots, \hat{y}^{(q)}, y) = \|\hat{y}-y\|_2^2$. 

\begin{Definition}
We define the  neural tangent kernel of a $C^2$ learning function $\mathcal{N}$ with $q$ outputs by

$$\forall (x, \tilde{x}) \in \left(\mathbb{R}^p\right)^2, \textbf{K}_{\text{NTK}}(x, \tilde{x}) = J_{\mathcal{N}}(x;w) J_{\mathcal{N}}(\tilde{x};w)^T \in \mathbb{R}^{q \times q},$$
where $J_{\mathcal{N}}(x;w) = \left(\frac{\partial \mathcal{N}_i}{\partial w_j}(x;w)\right)_{(i,j) \in [|1,q|] \times [|1,d|]} \in \mathbb{R}^{q \times d}$ is the Jacobian of $w \mapsto \mathcal{N}(x; w)$ at $(x,w)$ for any $x \in \mathbb{R}^p$.

\noindent
Given the learning path $\bw$ and the iteration numbers $k \in [|0, K|]$, the discrete tangent kernel path is defined by
$$\forall (x, \tilde{x}) \in \left(\mathbb{R}^p\right)^2, \textbf{K}_{\text{NTK}}(x,\tilde{x};k) = J_{\mathcal{N}}(x; w(k)) J_{\mathcal{N}}(\tilde{x}; w(k))^T.$$

\end{Definition}

\begin{Theorem}
Consider a $C^2$ learning machine model $\mathcal{N}$ with $q$ outputs and the path of parameters $\bw$  learned from a training set $\mathcal{D}$ by discrete gradient descent with steps $(\eta_k)_{k \in [|0,K-1|]}$ for a loss function $\mathcal{L}$. Then $\mathcal{N}$ satisfies
$$\forall x \in \mathbb{R}^p, \mathcal{N}(x;w(K)) = \mathcal{N}(x;w(0)) - \frac{1}{N} \sum_{n=1}^N \sum_{k=0}^{K-1} \eta_k  \boldsymbol{K}_{\text{NTK}}(x,x_n;k) \boldsymbol{L}(k,n) + O\left(\sum_{k=0}^{K-1} \eta_k^2\right),$$
where $\textbf{L}(k,n) := \begin{pmatrix} \frac{\partial L}{\partial \hat{y}^{(1)}}(\hat{y}^{(1)}_n(w(k)), y_n) \hdots \frac{\partial L}{\partial \hat{y}^{(q)}}(\hat{y}^{(q)}_n(w(k)), y_n) \end{pmatrix}^T \in \mathbb{R}^M$.

\end{Theorem}

\noindent
\textbf{Proof}

\noindent
Like for the one-dimensional case, we can write the Taylor approximation of $w \in \mathbb{R}^d \mapsto \mathcal{N}(x;w) \in \mathbb{R}^q$ for any $x \in \mathbb{R}^p$,
$$\mathcal{N}(x; w(k+1)) = \mathcal{N}(x;w(k)) + J_{\mathcal{N}}(x;w(k)) (w(k+1)-w(k)) + O(\|w(k+1)-w(k)\|_2^2).$$
Using the formula of the gradient descent of Equation \eqref{eq:discrete_dyn_grad_desc} we obtain
$$\mathcal{N}(x; w(k+1)) = \mathcal{N}(x;w(k)) + J_{\mathcal{N}}(x;w(k)) \left(-\eta_k \nabla_w \mathcal{L}(w(k))\right) + O(\eta_k^2).$$
The chain rule applied on the gradient of $\mathcal{L}$ is also similar to the one-dimensional case, but has to take into account the $q$ outputs. To do so, we introduce $\hat{y}^{(m)}_n: w \in \mathbb{R}^d \mapsto \mathcal{N}_m(x_n;w(k)) \in \mathbb{R}$ for all $n \in [|1,N|]$ and $m \in [|1,q|]$, and get
$$\frac{\partial \mathcal{L}}{\partial w_j}(w(k)) = \frac{1}{N} \sum_{n=1}^N \sum_{m=1}^q \frac{\partial \mathcal{N}_m}{\partial w_j}(x_n;w(k)) \frac{\partial L}{\partial \hat{y}^{(m)}}(\hat{y}^{(m)}_n(w(k)), y_n).$$
This equation combined with the Taylor expansion yields
\begin{align*}
    \mathcal{N}(x;w(k+1)) = \mathcal{N}(x;w(k)) - \eta_k \frac{1}{N} \sum_{n=1}^N \sum_{m=1}^M \frac{\partial L}{\partial \hat{y}^{(m)}}(\hat{y}^{(m)}_n(k), y_n) J_{\mathcal{N}}(x;w(k)) \nabla_w \mathcal{N}_m(x_n; w(k)) + O(\eta^2_k),
\end{align*}
which can be rewritten as
\begin{align*}
    \mathcal{N}(x;w(k+1)) = \mathcal{N}(x;w(k)) - \eta_k \frac{1}{N} \sum_{n=1}^N  J_{\mathcal{N}}(x;w(k)) J_{\mathcal{N}}(x_n; w(k))^T \textbf{L}(k,n) + O(\eta^2_k),
\end{align*}
where $\textbf{L}(k,n) := \begin{pmatrix} \frac{\partial L}{\partial \hat{y}^{(1)}}(\hat{y}^{(1)}_n(w(k)), y_n) \hdots \frac{\partial L}{\partial \hat{y}^{(q)}}(\hat{y}^{(q)}_n(w(k)), y_n) \end{pmatrix}^T \in \mathbb{R}^q$.
Finally, a telescopic summation gives
$$\mathcal{N}(x;w(K)) = \mathcal{N}(x;w(0)) - \frac{1}{N} \sum_{n=1}^N \sum_{k=0}^{K-1} \eta_k \textbf{K}_{\text{NTK}}(x,x_n;k) \textbf{L}(k,n) + O\left(\sum_{k=0}^{K-1} \eta^2_k\right).$$

\section{Two elementary examples with explicit neural tangent kernel}

\subsection{Linear regression} \label{sec:specialcase-linearreg}
Linear regression is one of the simplest possible learning problems. Consider a training data set 
$\mathcal{D} = (x_n, y_n)_{n \in [|1,N|]} \in \left(\mathbb{R}^p \times \mathbb{R}\right)^N$ and  the linear learning function
\begin{align*}
    \mathcal{N}: \mathbb{R}^p \times \mathbb{R}^p \colon &\to \mathbb{R} \\
    (x, a) &\mapsto \langle a, x \rangle_{\mathbb{R}^p}.
\end{align*}
We are here in the special case where $d=p$.
The goal of this function is to learn $a \in \mathbb{R}^p$  minimizing $\mathcal{L}(a) = \frac{1}{N} \sum_{n=1}^N (\langle a, x_n \rangle - y_n)^2$. The NTK of this learning machine simply is
$$K_{\text{NTK}}(x, \tilde{x}) = \langle x, \tilde{x} \rangle_{\mathbb{R}^p},$$
which is independent of the parameters $w := a$. Expanding the RHS of Equation \eqref{eq:domingo_discrete} using Equation \eqref{eq:discrete_dyn_grad_desc} gives
$$\forall x \in \mathbb{R}^n, \mathcal{N}(x; w(K)) = \mathcal{N}(x; w(0)) - \frac{1}{N} \sum_{n=1}^N \sum_{k=0}^{K-1} \eta_k \frac{\partial L}{\partial \hat{y}}(\hat{y}_n(k), y_n) K_{\text{NTK}} (x, x_n; k).$$
Hence, the $O( \cdot )$ term disappears and the formula becomes exact.
The neural tangent kernel is a polynomial kernel of degree one. One can note that in this case the normalized version of the NTK defined by
$$\frac{K_{\text{NTK}}(x, \tilde{x})}{\sqrt{K_{\text{NTK}}(x, x) \cdot K_{\text{NTK}}(\tilde{x}, \tilde{x})}},$$
is the cosine similarity.

\subsection{Neural networks with a linear last layer} \label{sec:theo-svm}
Given a network $\mathcal{N}: \mathbb{R}^p \times \mathbb{R}^d \to \mathbb{R}$ and a set of weights $w \in \mathbb{R}^d$, we consider the classic case where the last layer is linear. So we  assume that there exists $(W, b) \in \mathbb{R}^r \times \mathbb{R}$ and a network $\mathcal{P}: \mathbb{R}^p \times \mathbb{R}^{d-r-1} \to \mathbb{R}^r$ such that
$$\forall x \in \mathbb{R}^p, \mathcal{N}(x; w) = W \cdot \mathcal{P}(x; \tilde{w}) + b,$$
where  $\tilde{w} \in \mathbb{R}^{d-r-1}$ is the set of weights $w \in \mathbb{R}^{d}$ deprived of $W$ and $b$. It follows that, for $x \in \mathbb{R}^p$,
$$\nabla_W \mathcal{N}(x;w) = \mathcal{P}(x; \tilde{w}), \hspace{2cm} \nabla_b \mathcal{N}(x; w) = 1.$$
Now, since the feature vector induced by the NTK of $\mathcal{N}$ is given by $\phi = \nabla_w \mathcal{N}$ (Definition \ref{def:feature-space}), we can rewrite it as $\phi = \begin{pmatrix}\nabla_W \mathcal{N}, \nabla_b \mathcal{N}, \nabla_w \mathcal{P}\end{pmatrix}$. Then, if we let $a=\begin{pmatrix}W, b, 0, \hdots, 0\end{pmatrix} \in \mathbb{R}^d$, we obtain
$$\langle \phi(x), a \rangle = \mathcal{N}(x;w).$$
This means that the set of all the linear regressions that can be computed on the feature space \textit{contains the network itself}.
As an example, consider the case where a network is trained on a classification task where the goal is to predict a positive value for positive samples and a negative value for negative samples. If the network has a 100\% accuracy, then the feature space will be perfectly linearly separable.


\section{Numerical verification of Domingos' formula on two more examples}\label{sec:numverif-domingo}

In this section, we consider a  simple and controlled setting where the RHS of Equation \eqref{eq:domingo_discrete} can be effectively computed. First, we use this setting to measure how well the RHS approximates the LHS when varying different hyper-parameters.

In Domingos' proof (Section \ref{sec:domingo-proof}) and in its discrete formulation (Section \ref{sec:discrete-domingo-proof}), a few details differ from usage in network training. The main difference lies in the optimization being done using a full-batch gradient descent algorithm. In practice, each iteration optimizes the loss on a randomized mini-batch, and the optimization algorithms are more complex than simple gradient descent (for example by including ``momentum'' terms, or using a different step for each coordinate). For simplicity, we nevertheless consider a full-batch gradient descent in this section, like in Domingos' setting.

\subsection{First experiment: a ball vs sphere classification problem} \label{sec:sphere_exp}
\begin{figure}[H]
    \centering
    \begin{subfigure}{0.33\textwidth}
        \centering
        \includegraphics[width=\linewidth]{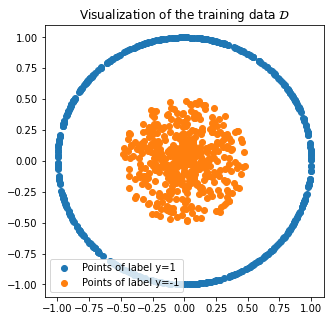}
    \end{subfigure}
    \begin{subfigure}{0.33\textwidth}
        \centering
        \includegraphics[width=\linewidth]{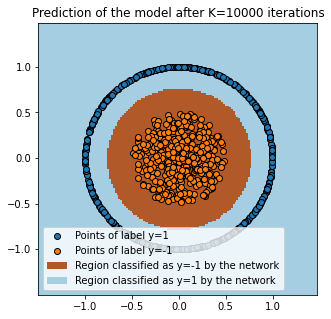}
    \end{subfigure}
    \caption{On the left, the points of the training dataset $\mathcal{D}$ for the first experiment. On the right, the prediction of the network after $K=10^4$ training iterations.}
    \label{fig:dataset_results_sphere}
\end{figure}

For this experiment, the dataset we considered was $\mathcal{D}$ such that:
$$\mathcal{D} \subset \{ (x, 1) : x \in \mathbb{R}^2, \|x\| = 1 \} \cup \{ (x, -1) : x \in \mathbb{R}^2, \|x\| \leq 0.5 \} \subset \mathbb{R}^2 \times \{ -1, 1 \}.$$
We consider here the case $p=2$ and $q=1$. The set we considered had $N=2^{10}$ different samples. See Figure \ref{fig:dataset_results_sphere} for an illustration.
We used the 3-layers neural network 
$$\forall x \in \mathbb{R}^p, \mathcal{N}(x;w) := W_1 \cdot \textsc{GELU}(W_2 \cdot \textsc{GELU}(W_3 \cdot x + b_3) + b_2) + b_1,$$
where $W_1 \in \mathbb{R}^{1 \times r}$, $W_2 \in \mathbb{R}^{r \times r}$, $W_3 \in \mathbb{R}^{r \times p}$, $b_1 \in \mathbb{R}^1$, $b_2 \in \mathbb{R}^{r}$, $b_3 \in \mathbb{R}^r$ and $w = (W_1, W_2, W_3, b_1, b_2, b_3)$. The  $C\infty$ $\textsc{GELU}$ function  is defined by
$$\forall s \in \mathbb{R}, \textsc{GELU}(s) = s \cdot \mathbb{P}(X \leq s),$$
where $X \sim \mathcal{N}(0, 1)$. It  is applied component-wise on a vector. The network has a total of $d = r^2 + (p + 3)r + 1$ parameters. To efficiently compute the NTK in the RHS of Equation \eqref{eq:domingo_discrete}, we derived manually the gradients of the network w.r.t the input and computed the scalar product.
Here $p=2$, and the network was trained for $K=10^4$ iterations using a constant learning rate $\eta = 10^{-2}$ with $N=2^{10}$ training samples. The loss  was the mean squared error (MSE) and the network's width, namely the size of each layer, was $r=10$.  

\subsubsection{Results}
The network quickly reached an accuracy of 100\%, meaning that each point of $\mathcal{D}$ was classified correctly. See Figure \ref{fig:dataset_results_sphere} for an illustration of the result. One can see in this figure that the boundary of decision seems to be the ball of radius $0.75$, which is a fair trade-off between fitting the training data and overfitting. It is worth noting that in this particular case, the network seems to have converged to an optimal algorithm.

\subsubsection{Visualization of the NTK} \label{sec:algo_sphere_exp}
\begin{figure}[H]
    \centering
    \includegraphics[width=\linewidth]{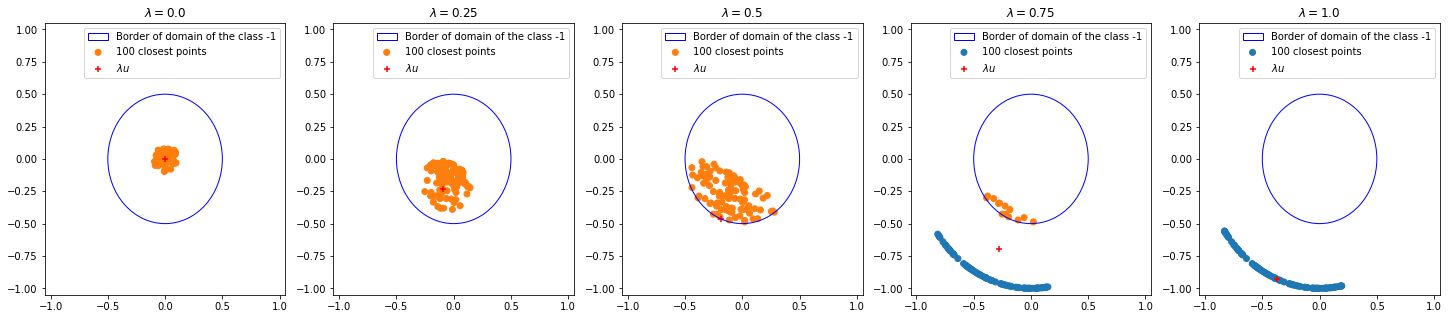}
    \includegraphics[width=\linewidth]{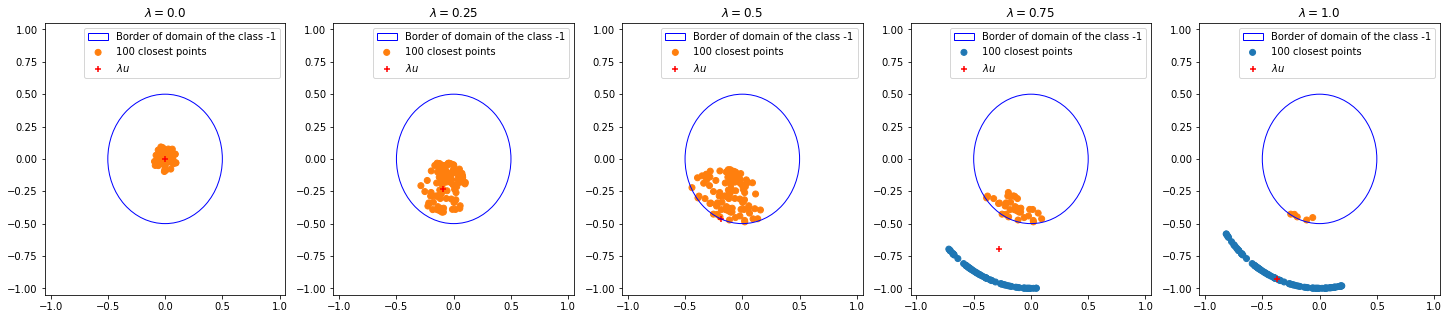}
    \caption{Results of the algorithm described in Section \ref{sec:algo_sphere_exp}. First row: 100 closest neighbors from the NTK's perspective (for $k=K-1$). Second row: 100 closest neighbors according to the Euclidean distance.}
    \label{fig:visualize_closest_sphere}
\end{figure}

To determine which points were considered as similar from the NTK's perspective, we computed the 100 points with largest similarity according to the normalized NTK. Namely, we considered the following algorithm:
\begin{itemize}
    \item[-] Draw a random direction of the space $u \in \mathbb{R}^2$ such that $\|u\| = 1$;
    \item[-] Compute the similarity vector $S(\lambda) = \left(\frac{K_{\text{NTK}}(\lambda u, x_1)}{\sqrt{K_{\text{NTK}}(\lambda u, \lambda u) \cdot K_{\text{NTK}} (x_1, x_1)}}, \hdots, \frac{K_{\text{NTK}}(\lambda u, x_N)}{\sqrt{K_{\text{NTK}}(\lambda u, \lambda u) \cdot K_{\text{NTK}} (x_N, x_N)}} \right)$ where the NTK is taken at $k=K-1$;
    \item[-] Compute the sequence $i_1(\lambda), \hdots i_N(\lambda)$ such that $S(\lambda)_{i_1(\lambda)} \geq \hdots \geq S(\lambda)_{i_N(\lambda)}$;
    \item[-] For different values of $\lambda$, display in a figure the points $S(\lambda)_{i_1(\lambda)}, \hdots, S(\lambda)_{i_{100}(\lambda)}$.
\end{itemize}
The result is available in Figure \ref{fig:visualize_closest_sphere}, along with the result of the same experiment but conducted using the Euclidean distance. We can see that both sets of neighbors are not identical, but similar. Therefore, the normalized NTK seems to be compatible with an Euclidean distance.

\subsubsection{Linear separability of the feature space defined by the NTK}
\begin{figure}[H]
    \centering
    \begin{subfigure}{0.33\textwidth}
        \centering
        \includegraphics[width=\linewidth]{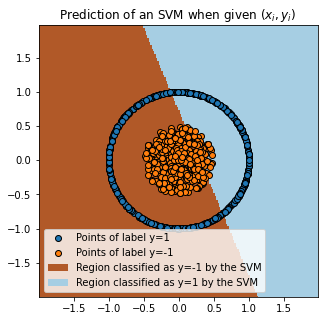}
    \end{subfigure}
    \begin{subfigure}{0.33\textwidth}
        \centering
        \includegraphics[width=\linewidth]{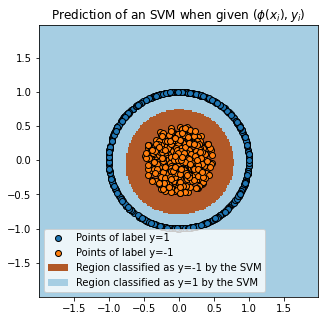}
    \end{subfigure}
    \caption{Results of the experiments on the linear separability of the feature space defined by the NTK. On the left are displayed the prediction of the SVM when trained on the \textbf{input space}. It fails On the right, the perfect prediction of the SVM when trained on the \textbf{NTK feature space}.}
    \label{fig:results_svm_sphere}
\end{figure}

The previous experiments suggest that the normalized NTK reflects the trained network's decisions. In this section, we empirically verify the claim of Section \ref{sec:theo-svm} stating that, since the network has a perfect accuracy, the feature space induced by the version of the NTK without normalization is linearly separable. As stated in Definition \ref{def:feature-space}, the feature vector induced by the NTK is defined by the transformation:
$$\phi: x \in \mathbb{R}^2 \mapsto \nabla_w \mathcal{N}(x; w) \in \mathbb{R}^d.$$ 
We trained an SVM \cite{cortes1995support} algorithm to output the label $y_i$ when given $\phi(x_i)$, for each input pair $(x_i, y_i) \in \mathcal{D}$. The results are available in Figure \ref{fig:results_svm_sphere}. We see in this figure that the decision boundary obtained by the SVM  closely matches the one obtained by the network, and that it reaches a 100\% accuracy. This indicates that the feature space induced by the NTK is linearly separable.

\subsection{Second experiment: the linearly separable case}
\begin{figure}[H]
    \centering
    \begin{subfigure}{0.33\textwidth}
        \centering
        \includegraphics[width=\linewidth]{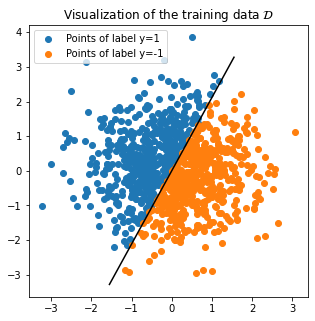}
    \end{subfigure}
    \begin{subfigure}{0.33\textwidth}
        \centering
        \includegraphics[width=\linewidth]{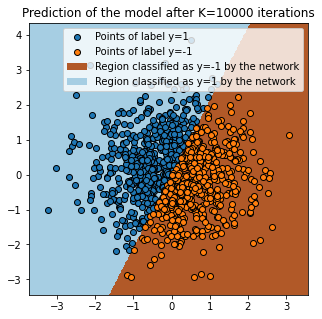}
    \end{subfigure}
    \caption{On the left are displayed the points of the training dataset $\mathcal{D}$ for the second experiment. On the right are the prediction of the network after $K=10^4$ training iterations.}
    \label{fig:dataset_results_linsep}
\end{figure}

\begin{figure}[H]
    \centering
    \includegraphics[width=\linewidth]{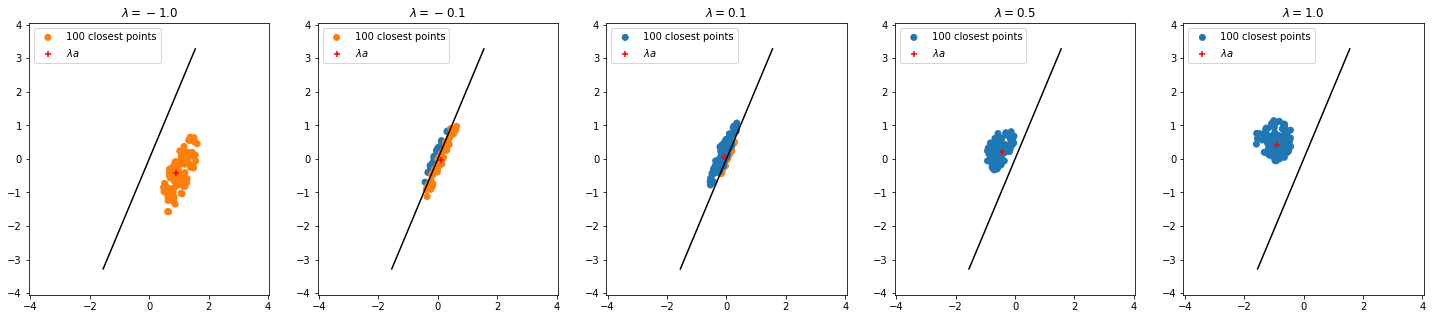}
    \includegraphics[width=\linewidth]{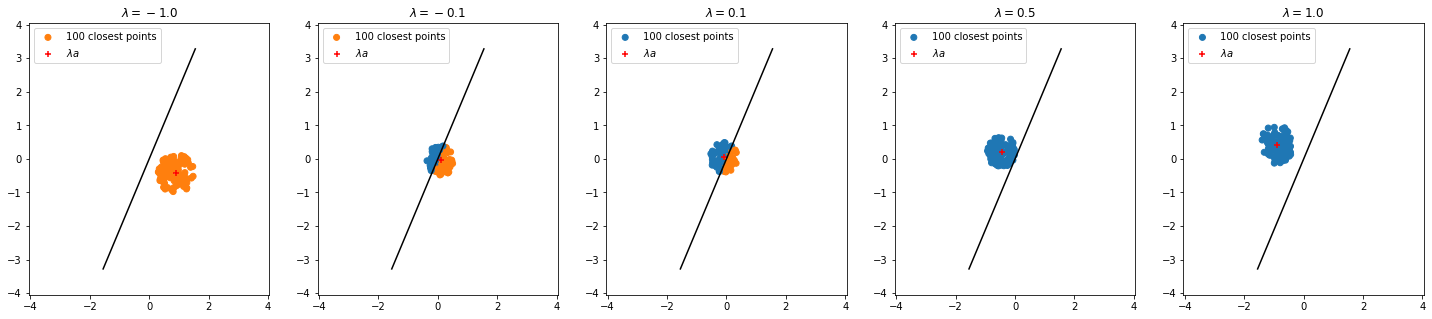}
    \caption{Results of the algorithm described in Section \ref{sec:sphere_exp}, but for the second experiment. First row: 100 closest neighbors from the NTK's perspective (for $k=K-1$). Second row: 100 closest neighbors according to the Euclidean distance.}
    \label{fig:visualize_closest_linsep}
\end{figure}

For the next toy experiment, we considered the same network and training settings as for Section \ref{sec:sphere_exp}. The dataset, on the other hand, was different and defined by
$$\mathcal{D} \subset \{ (x, 1) : x \in \mathbb{R}^2, \langle x, a \rangle \geq 0 \} \cup \{ (x, -1) : x \in \mathbb{R}^2, \langle x, a \rangle < 0 \} \subset \mathbb{R}^2 \times \{ -1, 1 \}.$$
Like for the first experiment, we considered $N=2^{10}$ training data points. The accuracy obtained by the network after training was $99.80\%$. The dataset and the prediction of the trained network are illustrated in Figure \ref{fig:dataset_results_linsep}. We display in Figure \ref{fig:visualize_closest_linsep} the results of the algorithm described in Section \ref{sec:sphere_exp} where we considered $u = a$. We see that here again the network learns a deformation of the input space such that most of the closest neighbors share the same label. Furthermore, one can note that the learned deformation follows the direction orthogonal to $a$ and is localized around the center of the space.

\section{Can a network learn planar topology?}\label{sec:shape-completion-2d}

Returning to a fruitful example in the discussion of the performance of neural networks, we reconsider the problem of learning connectivity of planar shapes, a problem that was already at the core of the debate on the perceptron's performance in the seventies of the past century \cite{minsky1969perceptrons}.
\begin{figure}[H]
    \centering
    \begin{subfigure}{.15\textwidth}
        \centering
        \includegraphics[width=\linewidth]{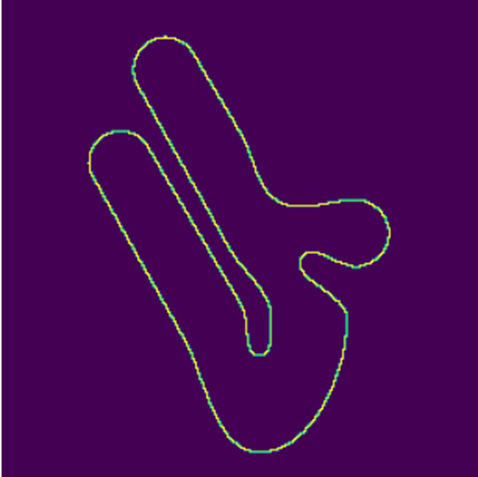}
    \end{subfigure}
    \begin{subfigure}{.15\textwidth}
        \centering
        \includegraphics[width=\linewidth]{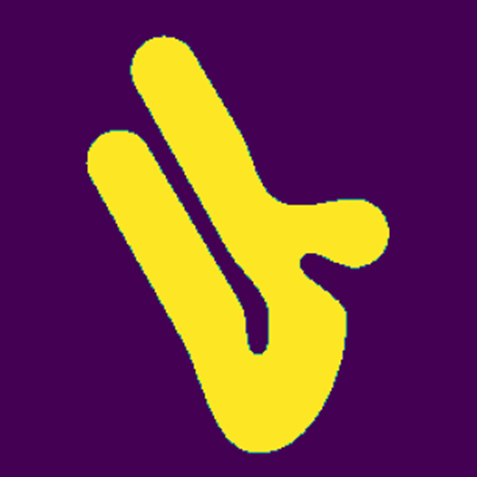}
    \end{subfigure}
    \\
    \begin{subfigure}{.15\textwidth}
        \centering
        \includegraphics[width=\linewidth]{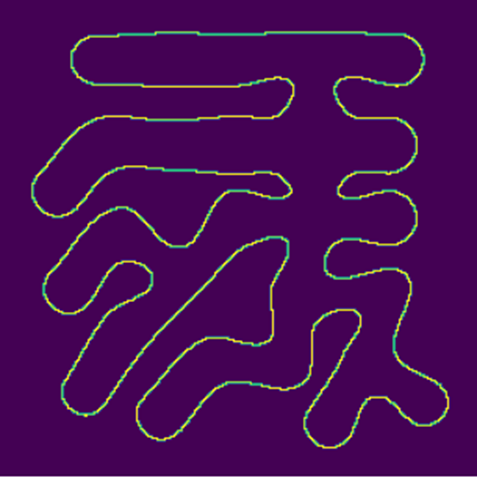}
    \end{subfigure}
    \begin{subfigure}{.15\textwidth}
        \centering
        \includegraphics[width=\linewidth]{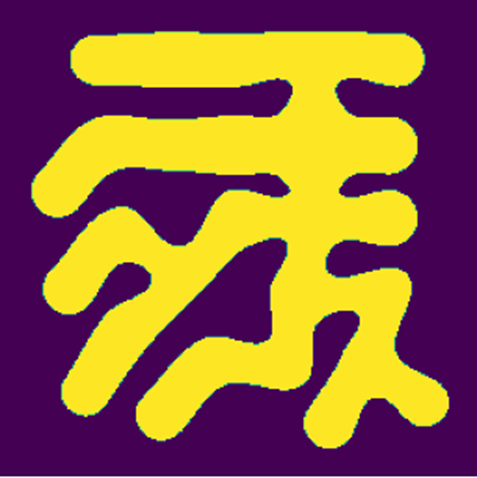}
    \end{subfigure}
    \caption{Examples of data samples (left column) and their associated label (right column) for the shape completion dataset. The shapes are not convex making it impossible for a network to solve the task based only on local cues.}
    \label{fig:examples_sample}
\end{figure}
The generic question behind is whether a network can learn from examples the optimal algorithm relative to a given task, provided this optimal algorithm is actually compatible with the architecture of the network. In other terms, if its internal right parameters $w$ were correctly chosen, this  network would  perform exactly what is being asked. But this does not imply that training it by gradient descent will reach this perfect configuration. This is all the point of Domingo's theorem.

We have seen in the simple examples presented in Section \ref{sec:numverif-domingo} that the network can sometimes converge to the perfect configuration. But does this property remain when the considered data is high-dimensional?
Domingos' theorem, as we saw, argues to the contrary, namely a network optimized by gradient descent is inherently unable to perform but an interpolation of its learning data. There is no argument showing that it could be  able of another more efficient form of interpolation. 
To verify this, our learning dataset was made of connected shapes not meeting the image boundary. The input to the network was the border of this shape, and we trained the network to output the filled in shape. See Figure \ref{fig:examples_sample} for an illustration.

An elementary low complexity algorithm  solves this task. It is enough  to browse the pixels of the image from left to right and to attribute to the corresponding pixel of the output image the congruence modulo 2 of the sum of the previously seen pixels. In other words, for an image of $\{0,1\}^{H \times W}$ with height $H$ and width $W$, the algorithm can be written as:
\begin{equation}
    \label{eq:algo}
    \forall (i, j) \in [|1,H|] \times [|1,W|], y_{i,j} = \sum_{k=1}^j x_{i,j} \mod 2.
\end{equation}
It is an easy exercise to check that this algorithm is implementable in any deep neural network with a sufficient number of layers and we shall present and study one of such architecture in Section \ref{sec:1dplanar}.

\begin{figure}[H]
    \centering
    \begin{subfigure}{.15\textwidth}
        \centering
        \includegraphics[width=\linewidth]{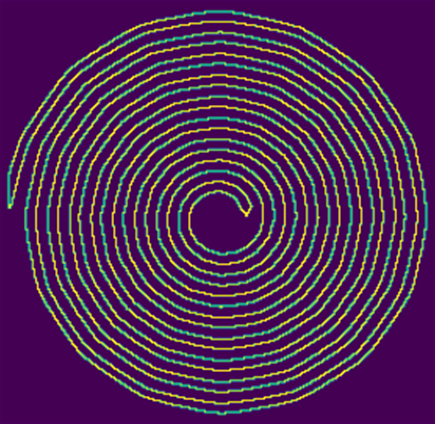}
    \end{subfigure}
    \begin{subfigure}{.15\textwidth}
        \centering
        \includegraphics[width=\linewidth]{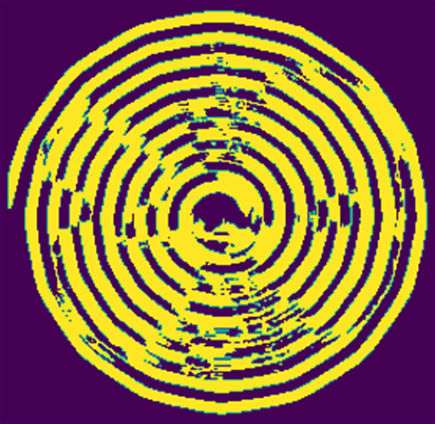}
    \end{subfigure}
    \caption{Example of a more complex case where the network fails. If the network had learned an optimal algorithm, it would have managed to deal with this case. For this example, the IOU is 51\%.}
    \label{fig:hard_example}
\end{figure}

Then, we trained a U-Net on a large database of connected shapes and we found that it reached an almost 100\% accuracy. But, when presented with samples differing too much from the ones seen during training, the network failed (Figure \ref{fig:hard_example}).

\subsection{Study of a structure which can learn 1D planar topology} \label{sec:1dplanar}
As mentioned in the previous section, there is an elementary 1D algorithm solving the 2D shape filling problem. To understand better the failure to extrapolate to new data pointed out in the previous section, we explored the 1D shape completion problem.
\subsubsection{Presentation of the dataset}
The problem of learning planar topology in one dimension consists, given a sample of $\{0,1\}^p$, in applying the algorithm of Equation \eqref{eq:algo} to this sample. It is possible to learn independently the task of Figure \ref{fig:examples_sample}  for each line of the image. To better control the difficulty of the task, we can constrain the number of nonzero coordinates. To this end, we introduce
$$\forall l \in \{ k \in [|1, p|] : k \text{ is even}\}, \mathcal{X}_l = \left\{ x \in \{0,1\}^p : \sum_{i=1}^p x_i = k \right\}$$
and for each sample of $\mathcal{X}_l$, we define  its output (or label) is defined as the result of Equation \eqref{eq:algo} on it. It follows that the number of outputs is $q=p$. We can define the training dataset as
$$\mathcal{D}_l \subset \mathcal{X}_l \times \{0,1\}^p$$
In particular, one can note that $|\mathcal{X}_l| = \binom{p}{l}$ and that the number of possibilities for $p = 128$ and $l=10$ is of order $10^{14}$. For the experiments in this section, we considered $N=10^4$ training samples and $p=128$.

\subsubsection{An NN structure that can solve the problem}
This structure is based on the following observation: 
$$\forall (u, v) \in \{0,1\}^2, u + v \mod 2 = \max(0, u-v) + \max(0, v-u).$$
In particular, $s \in \mathbb{R}, \mapsto \max(0, s) = \textsc{ReLU}(s)$, an activation function commonly used in neural networks. The operation described above is hence implementable using two linear operations and a non-linearity, \textit{i.e.}
$$\forall (u, v) \in \{0,1\}^2, u + v \mod 2 = \begin{bmatrix}1 & 1\end{bmatrix} \cdot \textsc{ReLU}\left(\begin{bmatrix}1 & -1 \\ -1 & 1\end{bmatrix} \cdot \begin{bmatrix}u \\ v\end{bmatrix}\right).$$
Furthermore, the above operation can be implemented as a convolution if we want to compute the sum modulo two of all the pairs of consecutive samples in a point $x \in \{0,1\}^p$. We shall denote $\Lambda: \{0,1\}^p \to \{0,1\}^{p-1}$ this operation.

\begin{figure}
    \centering
    \includegraphics[width=0.5\linewidth]{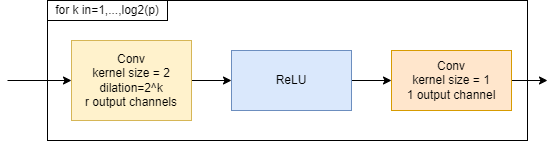}
    \caption{Description of the structure of a neural network that can perfectly solve the one dimensional planar topology problem. The smallest $r$ for which it is still possible to solve the task with this structure is $r=2$.}
    \label{fig:optimal-structure}
\end{figure}

Implementing Equation \eqref{eq:algo} also requires computing the cumulative sum. To this end, we shall restrict the analysis to the case where $p$ is a power of 2,  and we consider $\Gamma: x \in \{0,1\}^p \mapsto \Lambda(x) \downarrow \ \in \{0,1\}^{p/2}$ where $\downarrow$ corresponds to the downsampling operation, which consists in taking one sample out of two, starting with the first one. Then, we call $\Gamma^k$ the  $k$ times composition of the $\Gamma$ function and remark that 
$$\Gamma^k(x)_0 = \sum_{i = 1}^{2^k} x_i \mod 2.$$
Finally, we note $\Delta(x;k) = (0,\hdots,0, x_1, \hdots, x_k) \in \{0,1\}^p$ the function that takes the first $k$ coordinates of a sample $x \in \{0,1\}^p$ and zero-pads it on the left so as to obtain a vector of the same dimension. Then
$$\begin{pmatrix}\Gamma^{\log_2(p)}(\Delta(x;1))_0 & \hdots & \Gamma^{\log_2(p)}(\Delta(x;p))_0\end{pmatrix} \in \{0,1\}^p$$
is precisely the result of the application of Equation \eqref{eq:algo} to a sample $x \in \{0,1\}^p$. To conclude, we remark that every operation we described above are compatible with a convolutional structure: the above formula describes a neural network with $2 \cdot \log_2(p)$ convolutions, $\log_2(p)$ non-linearities and $\log_2(p)$ downsamplings. The resulting structure can be seen in Figure \ref{fig:optimal-structure}.

\subsubsection{Training the structure}
When training the structure described in Figure \ref{fig:optimal-structure} with $r < 8$ and for any number nonzero coordinates $l$ and despite heavy hyper-parameter tuning, the network never converged to the optimal solution. An exception to this observation  was obtained when the initialization of the network was  $w^* + \sigma \mathcal{N}(0,1)$ where $\sigma < 0.1$ and where $w^*$ was the manually-computed optimal solution.

For any $r \geq 8$, the network converged without any further help to an optimal solution for any number of nonzero coordinates $l$. For instance, when training for $l=10$ and evaluating on $10^5$ samples of $\mathcal{D}_{20}$, we reached an error of $10^{-7}$. Thus in that very particular case, the network learned the correct algorithm and proved able of ``generalization''.

When changing slightly the architecture though, for instance by increasing the kernel size of one of the convolutions, the network stopped converging for any $r$ and any $l$.

When significantly increasing the number of parameters by the introduction of other convolutions in-between the two presented in Figure \ref{fig:optimal-structure}, the network converged to a solution that is not the optimal one, but whose error was low. As an example, when training for $l=10$, we obtained an error of order $10^{-4}$ when evaluated on $\mathcal{D}_4$, $10^{-3}$ for $\mathcal{D}_{10}$ and $10^{-2}$ for $\mathcal{D}_{20}$.

All in all, this indicates that \textit{even when provided with an optimal structure, the network does not generally converge to the optimal solution, unless heavily guided.} Any change in the structure puts at risk this convergence. This example brings one more evidence in favor  of this interpretation of Domingos's theorem:  a network only learns based on similarity and there will never be a guarantee of perfect interpolation nor extrapolation.

\subsection{Extension to the 2D planar topology with an usual structure}

\subsubsection{Presentation of the dataset}
The shapes were generated using the code associated with an online demo \cite{shapegenerator} which creates SVG files with one connected shape. The SVG files were converted to JPEG images of dimension $512 \times 512$. At this point the images had binary values. Then, we computed the label associated with each image by computing the boundary of the shape as the set of pixels in the shape that have at least one of its eight neighbors outside of the shape. Finally, these images and their associated labels were filtered using a Lanczos filtering, downsampled by a factor of two and saved. This filtering created a small smoothing effect around the boundaries of the object. See Figure \ref{fig:examples_sample} for an example. Overall, 10,000 training images and 1,000 test images were generated this way.

\subsubsection{Experimental setup}
\begin{figure}[H]
    \centering
    \includegraphics[width=0.8\linewidth]{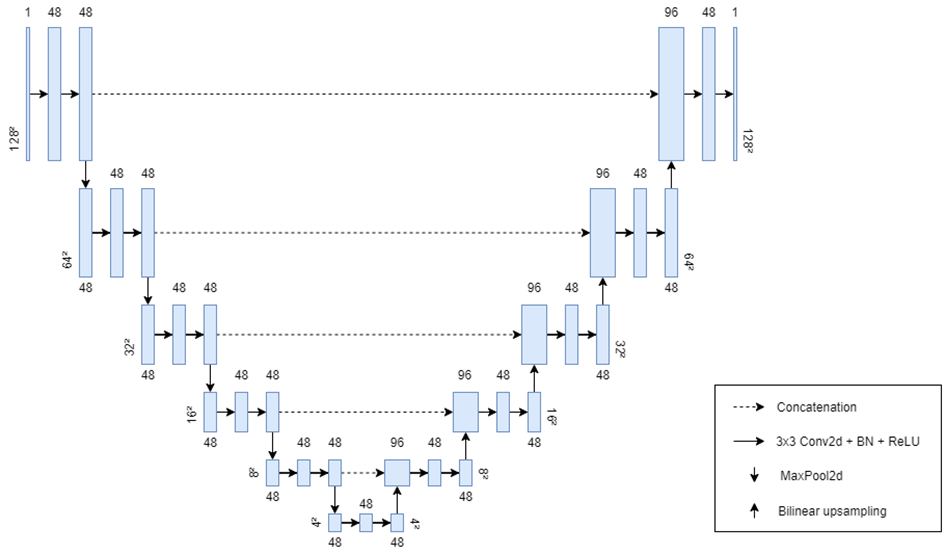}
    \caption{Description of the U-Net used for the experiments on the shape completion dataset. The U-Net processes the input at six different scales.}
    \label{fig:unet}
\end{figure}

We considered a U-Net architecture \cite{ronneberger2015u} (see Figure \ref{fig:unet}) which allows for a processing of the entire image due to its multiple down-sampling operations. Its hourglass structure and its multiple skip connections allow for a powerful yet easy-to-train model.

To speed up the training, the images were down-sampled by a factor of two before being passed through the network. Therefore, the images seen by the network were of dimension $128 \times 128$. The dataset was augmented by random horizontal and vertical flips.

The network was trained with the Ranger optimizer (a mix of \cite{liu2019variance} and \cite{zhang2019lookahead}) for 150 epochs using the mean square error loss and a batch size of 32. The learning rate was constant and equal to $10^{-2}$ and was divided by 10 whenever a plateau was reached \textit{i.e.} when the loss stopped decreasing for a few iterations.

\subsubsection{Results}
\begin{figure}[H]
    \centering
    \begin{subfigure}{.25\textwidth}
        \centering
        \includegraphics[width=\linewidth]{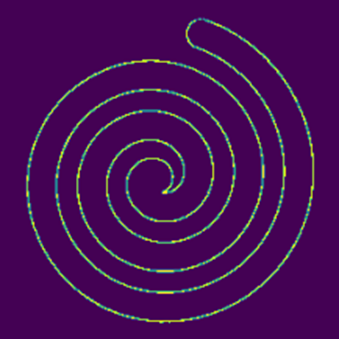}
    \end{subfigure}
    \begin{subfigure}{.25\textwidth}
        \centering
        \includegraphics[width=\linewidth]{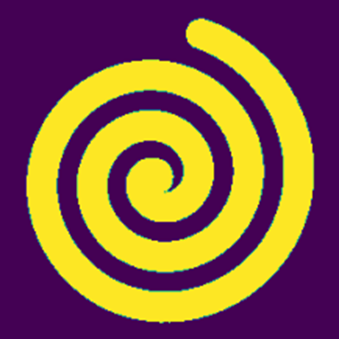}
    \end{subfigure}
    \\
    \begin{subfigure}{.25\linewidth}
        \centering
        \includegraphics[width=\linewidth]{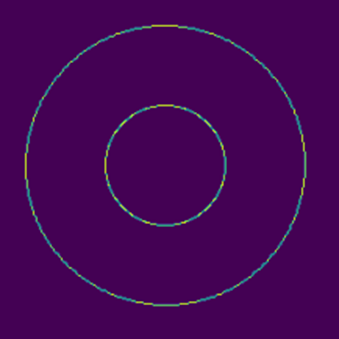}
    \end{subfigure}
    \begin{subfigure}{.25\linewidth}
        \centering
        \includegraphics[width=\linewidth]{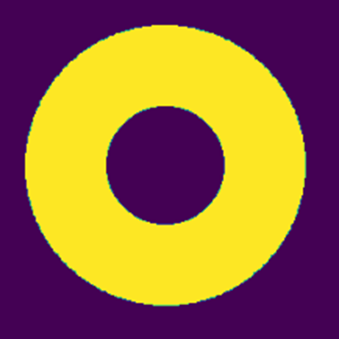}
    \end{subfigure}
    \caption{Examples of tricky cases that had not been seen during training and where the network  manages to make a perfect prediction. Notably, the first example could confuse a human observer. On the left, the input to the network. On the right, the prediction produced by the network.}
    \label{fig:tricky_examples}
\end{figure}

The network reached a final IOU (Intersection Over Union) of 99.80\%. When presented with complex examples that were not in the training set, the network still managed to perfectly fulfill its task. See Figure \ref{fig:tricky_examples} for a visual illustration.

This does not mean that the network learned the perfect algorithm. Indeed, further increasing the complexity of the presented shape causes the network to fail! See for instance Figure \ref{fig:hard_example}. If the network had learned an optimal algorithm, it would have managed to deal with such more complex cases. If, instead, as argued by Domingos, the network is guided by a comparison with known examples, there was not point in hoping success with an ``out of domain'' example.

\subsection{Using the NTK to better understand one-dimensional planar topology} \label{sec:shape1d}
As the formula of the NTK involves heavy computations and high-dimensional vectors, we were not able to compute it for the 2D problem described above. Instead, we considered a one-dimensional version of the problem that we describe in this section. This dataset consisted in sequences with only two non-zero values. This dimension reduction allowed us to work with smaller networks and thus make the computation of the NTK tractable. The dataset can be described by
$$\mathcal{D} \subset \mathcal{X} \times \{ 0, 1 \}, \hspace{1cm} \mathcal{X} := \left\{ x \in \{0,1\}^{p} : \exists (k,l) \in [|1,p|]^2, k \neq l, \forall n \in [|1,p|], \begin{cases}x_n = 1 & \text{ if } n \in \{ k, l \} \\ x_n = 0 & \text{ else}\end{cases} \right\},$$
where $p = 64$ in all our experiments. One may note that the input space $\mathcal{X}$ introduced here is the $\mathcal{X}_2$ described in Section \ref{sec:1dplanar}. What differed from Section \ref{sec:1dplanar} however is the way the label was computed. 

We call $\psi_1: \mathcal{X} \to [|1,p|]$, $\psi_2: \mathcal{X} \to [|1,p|]$ the indices such that for each sample $x \in \mathcal{X}$, we have $x_{\psi_1(x)} = x_{\psi_2(x)} = 1$ and $\psi_1(x) < \psi_2(x)$.

\begin{figure}[H]
    \centering
    \begin{subfigure}{0.33\textwidth}
        \centering
        \includegraphics[width=\linewidth]{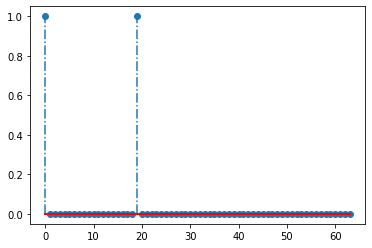}
    \end{subfigure}
    \begin{subfigure}{0.33\textwidth}
        \centering
        \includegraphics[width=\linewidth]{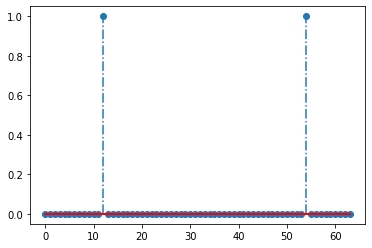}
    \end{subfigure}
    \caption{Examples of data used in the experiment of Section \ref{sec:shape1d}. On the left, the label is 0 because the point of index 32 does not lie between the two peaks. On the right, the label is 1.}
    \label{fig:dataset_shape1d_examples}
\end{figure}

The learning task was to find the value of the central pixel after shape completion. In other terms, for an input sample $(x_i, y_i) \in \mathcal{D}$, $y_i = \mathbf{1}_{\{\frac{p}{2} \in [|\psi_1(x_i), \psi_2(x_i)|]\}}$. Examples of inputs with different labels are shown in Figure \ref{fig:dataset_shape1d_examples}. Note that, although simple, this task cannot be resolved by looking at the closest neighbor algorithm in the input space: two points with the same Euclidean distance to a given point can have different labels. This is in opposition with the previous experiments where the NTK was shown to be closely related to the Euclidean distance.

In this experiment we also considered $N=2^{10}$ training data points. The trained network was the one of Section \ref{sec:sphere_exp} with $p=64$. We used a constant learning rate  $\eta=10^{-1}$ and $K=10^4$ training iterations. The network reached  100\% accuracy on both seen and unseen samples.

\begin{figure}[H]
    \centering
    \includegraphics[width=0.5\linewidth]{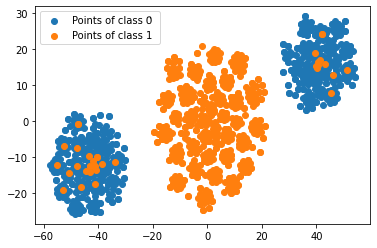}
    \caption{The t-SNE representation of the feature space learned by the network for one-dimensional shape completion problem. The algorithm is class-agnostic. The points shown are the aggregation of both the training and the test set.}
    \label{fig:tsne-shape1d}
\end{figure}

The data being high-dimensional could not be displayed in a normal plot. Instead, we computed a t-SNE \cite{van2008visualizing} representation of the feature space $(\phi(x_i))_{i \in [|1,N|]}$ (Def. \ref{def:feature-space}), see Figure \ref{fig:tsne-shape1d}. The t-SNE algorithm is  class-agnostic, therefore the class of each data point is not given to the algorithm. In this figure, one can observe how the data are split into three distinct locations of the 2D plane. The left-most location corresponds to data points $x_i$ where both $\psi_1(x_i) < \psi_2(x_i) \leq \frac{p}{2}$. Points of class one that are located on the left correspond to cases with $\psi_1(x_i) = \frac{p}{2}$. A similar property holds for data points $x_i$ on the right-most part of the plane: $\frac{p}{2} \leq \psi_1(x_i) < \psi_2(x_i)$. The points of class 1 located on the right are also points with $\psi_1(x_i) = \frac{p}{2}$. Finally, all the points lying in the middle of the plane correspond to points of class 1, \textit{i.e.} points $x_i$ where $\psi_1(x_i) \leq \frac{p}{2}$ and $\psi_2(x_i) \geq \frac{p}{2}$.

Lastly, we investigated the closest points with respect to the normalized NTK: $\frac{K_{\text{NTK}}(x, \tilde{x})}{\sqrt{K_{\text{NTK}}(x, x) K_{\text{NTK}}(\tilde{x}, \tilde{x})}}$ is the highest for data points $(x, \tilde{x}) \in \mathcal{X}^2$ such that $|\{ \psi_1(x), \psi_2(x) \} \cap \{ \psi_1(\tilde{x}), \psi_2(\tilde{x}) \}| = 1$, \textit{i.e.} data points that share exactly one peak. Apart from this observation, it is very hard to determine what the network does to separate classes based on the normalized NTK only. This is where the curse of dimensionality comes into play: the higher-dimensional the data, the more difficult it is to extract meaningful information from the NTK. 

All in all, our observations on the t-SNE and the normalized NTK seemed to indicate that the network had found an optimal algorithm. This was easy to check in our case, as $|\mathcal{X}| = \binom{p}{2} = 2,016$. Computing the accuracy across all $\mathcal{X}$ confirmed that the network had converged to an optimal algorithm.

While we would have liked to extend this result to data with  more than two peaks, it was practically impossible for this simple massive optimization (full batch SGD without momentum). All experiments featuring more than two peaks failed to maintain its level of precision outside the training set. Changing the optimizer for Adam \cite{kingma2014adam} fixed the this issue, but the algorithm being more complex, we could not derive an analogue of Theorem \ref{Theorem1} for this optimizer.

\section{Conclusion and limitations}
In this paper, we studied Domingos' proof stating that neural networks are approximately kernel methods. This study led us to examine the kernel that naturally arises when proving the statement, namely the Neural Tangent Kernel. We found that in a low-dimensional setting (small neural network, input of small dimension, one-dimensional output) this kernel provides a powerful insight as to how the network makes its predictions. In these settings, we noticed that the network converged to an optimal algorithm that interpolated and extrapolated perfectly.

However, when the dimension of the input grows, the interpretation grows more complex and it is unclear how the interpretability of the NTK would apply to high dimensional settings such as images, where even the nearest neighbors can look very different (see Figure 5 of \cite{charpiat2019input} for an example). 

These concerns are all the more justified since the gradient descent algorithm used throughout this paper is no longer usable to train neural networks in high dimension. When changing the optimizer for a more complex one (such as in Section \ref{sec:shape-completion-2d}), the networks seem to converge to an intermediate solution that interpolate well in the training domain. But when presented with significantly different samples, the predictions deteriorate. This seems to confirm Domingos' interpretation of his theorem, which states neural networks are little more than a sophisticated kernel machine, that simply interpolates between known data. 


{\small
\bibliographystyle{unsrt}
\bibliography{bibliography}
}

\end{document}